\pdfoutput=1

\documentclass[11pt]{article}

\usepackage{acl}

\usepackage{times}
\usepackage{latexsym}
\usepackage{booktabs}
\usepackage{graphicx}
\usepackage{tcolorbox}
\usepackage{adjustbox}
\usepackage{fancyvrb}
\usepackage{fvextra}
\usepackage{tikz}  
\usetikzlibrary{arrows.meta, positioning, shapes.geometric} 

\usepackage[T1]{fontenc}

\usepackage[utf8]{inputenc}

\usepackage{microtype}

\usepackage{inconsolata}

\usepackage{graphicx}

\usepackage{tcolorbox}

\newcommand{\accorig}{Acc_{Or}}
\newcommand{\acccorr}{Acc_{Co}}
\newcommand{\accclar}{Acc_{Cl}}

\title{Synthetic Clarification and Correction Dialogues about Data-Centric Tasks \\ A Teacher-Student Approach}

\author{Christian Poelitz, Nick McKenna \\
  Microsoft Research, Cambridge UK \\
  \texttt{cpoelitz@microsoft.com}}

\begin{document}
\maketitle
\begin{abstract}
Real dialogues with AI assistants for solving data-centric tasks often follow dynamic, unpredictable paths due to imperfect information provided by the user or in the data, which must be caught and handled. Developing datasets which capture such user-AI interactions is difficult and time-consuming.
In this work, we develop a novel framework for synthetically generating controlled, multi-turn conversations between a user and AI assistant for the task of table-based question answering, which can be generated from an existing dataset with fully specified table QA examples for any target domain. Each conversation aims to solve a table-based reasoning question through collaborative effort, modeling one of two real-world scenarios: (1) an AI-initiated clarification, or (2) a user-initiated correction. Critically, we employ a strong teacher LLM to verify the correctness of our synthetic conversations, ensuring high quality. We demonstrate synthetic datasets generated from TAT-QA and WikiTableQuestions as benchmarks of frontier LLMs. We find that even larger models struggle to effectively issuing clarification questions and accurately integrate user feedback for corrections. 

\end{abstract}

\section{Introduction}

\begin{figure}[t]
    \centering
    \includegraphics[width=0.9\linewidth]{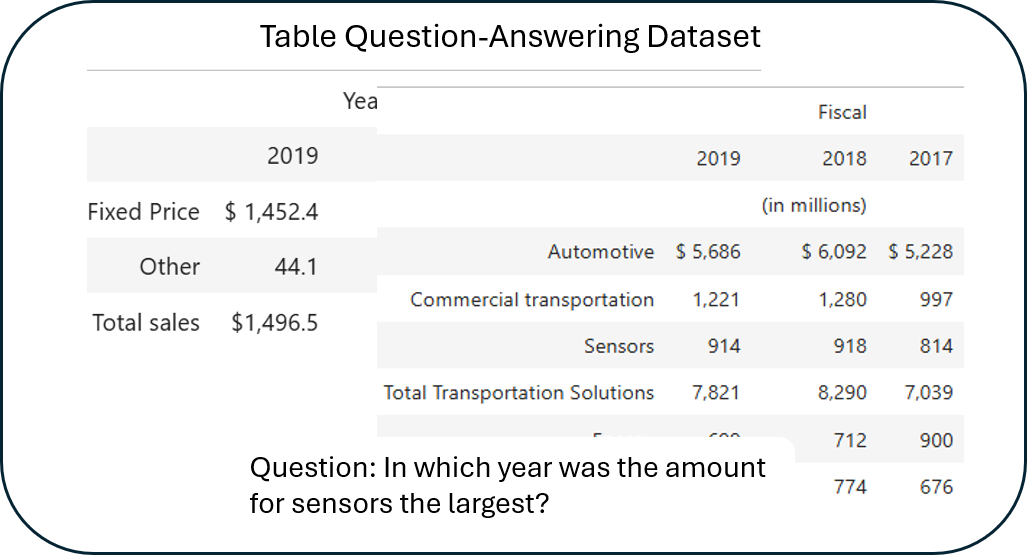}
    \includegraphics[width=1\linewidth]{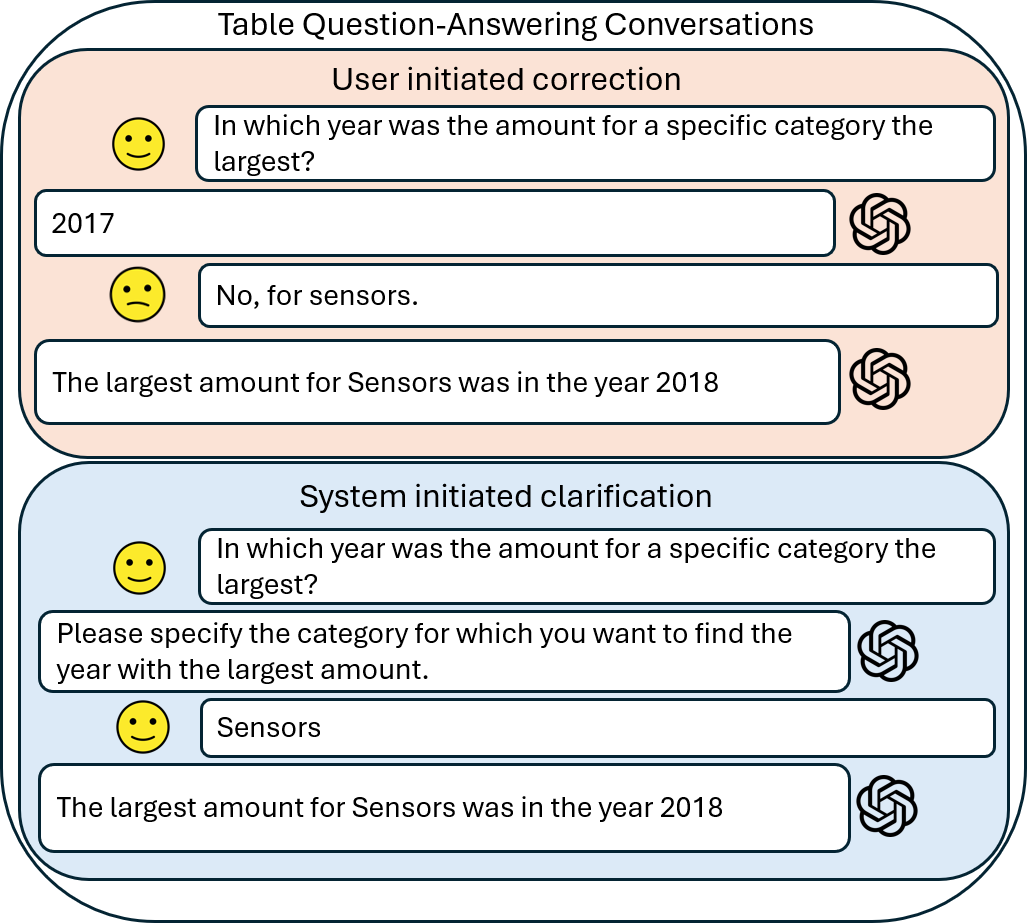}
    \caption{Illustration of synthetic corrections and clarifications based on table QA tasks from a dataset. Top: Original
    table QA dataset. Bottom: Two synthetic conversations with underspecified question. Top-bottom:
    User-initiated correction. Lower-bottom: AI-initiated clarification.}
    \label{fig:illustration}
\end{figure}

AI assistants are effective for solving data-centric tasks, in particular table-based question-answering (table QA)~\cite{Li_Sun_Cheng_2021, nararatwong-etal-2022-enhancing}, with numerical reasoning frequently addressed by generating code that operates on the table \cite{deng-etal-2022-pacific, chen2021finqa}. Yet, solving problems in dialogue with AI assistants is often more complex. These can be \textit{multi-turn} conversations with an unpredictable fashion, especially if the task depends on incomplete information both from a user and the data context to achieve a solution. Identifying missing information and recovering from any errors that arise is critical. To measure and improve the performance on such conversations we rely on available conversational data with relevant dialog acts of clarifications and corrections. Creating this dialogue data is often a laborious and manual task, making it difficult to reproduce for a target domain.

To address these gaps, we propose a novel framework with a \textbf{student-teacher} method for synthetically generating controlled, multi-turn conversations between a user and an AI assistant for table-based QA tasks, involving clarifications and corrections. Our approach leverages existing datasets from any target domain to create realistic synthetic dialogues that model real-world interactions. Each conversation is designed to solve a table-based reasoning question through collaborative effort, modeling one of two common scenarios:

\begin{enumerate}
    \item \textbf{AI-initiated clarification}: The assistant identifies ambiguous or incomplete information in the user's \textit{question} or \textit{data} and proactively asks for clarification to answer the question.
    \item \textbf{User-initiated correction}: The assistant provides an initial answer that may be incorrect due to missing information and a user provides the missing information as correction.
\end{enumerate}

In our approach, a teacher model collaborates with a student model (acting as the AI assistant) to generate synthetic multi-turn conversations. These dialogues, which incorporate both clarifications and corrections, form a curriculum for benchmarking (and training) the student model. On the curricula, we test how well the students are able to generate clarification questions without knowing whether the task can be solved with the available information, as well as how effective they can leverage user feedback to correct a wrong answer.

Our synthetic curriculum is designed with solvability as a core requirement. In every dialogue, we ensure that (i) the student model is capable, in principle, of asking the right clarification question, meaning tasks are never impossible; (ii) when the student poses a clarification question, there is a clear pathway to correct a wrong answer; and (iii) even if the student does not ask for clarification, simulated user interventions provide the necessary correction to eventually guide the model 


We make the following contributions:
\begin{enumerate}
    \item We present a novel student-teacher framework for generating realistic conversational data from arbitrary, existing data-centric tasks, offering a solution to the unpredictability and manual creation issues of current datasets.
    \item We establish baselines for several large language models (LLMs) on table QA benchmarks. In scenario 1, which involves model-initiated clarifications, we observe that even strong LLMs like GPT4-Turbo struggle with clarifications. In scenario 2, which involves user-initiated corrections, we find that on average the models perform better when they are directly corrected by a user rather than when they ask for clarification.
    \item We demonstrate the effectiveness of using additional synthetic generations as training data, resulting in significant improvements in model performance and robustness, increasing for example the accuracy of smaller LLMs like Llama3.1 8b and Qwen2.5 7b on clarifications.
    \item We will make all synthetically generated conversations and all associated code publicly available after publication.
\end{enumerate}

\section{Background and related work}
Our approach for synthetically generating conversations with clarifications and corrections in table QA intersects with several research areas, ranging from generating synthetic training data, synthetic benchmarks for LLMs and synthetic data for task specific dialogues.

\paragraph{Synthetic data generation}
Early approaches to generate synthetic data for training originate from traditional instruction tuning data~\cite{longpre2023flancollectiondesigningdata} and include approaches like self-instruct to generate instructions using LLMs~\cite{wang-etal-2023-self-instruct} or LLM-based augmentations with rewrites~\cite{xu2023wizardlmempoweringlargelanguage}. More recently, task specific synthetic train data using domain ontology have emerged~\cite{sudalairaj2024lablargescalealignmentchatbots, li2024syntheticdataalmostscratch}. For a more general overview on recent approaches for synthetic training data, see~\cite{long-etal-2024-llms, liu2024bestpracticeslessonslearned}.

Besides using LLMs to synthetically generate training data, recently more and more benchmarks for evaluations and automatic evaluation methods emerged leverage LLMs to generate data and evaluate model responses. Previous benchmarks~\cite{zheng2023judgingllmasajudgemtbenchchatbot, dubois2024lengthcontrolledalpacaevalsimpleway, lin2024wildbenchbenchmarkingllmschallenging,pmlr-v235-gu24c} mainly leverage synthetic data and LLMs for evaluations. Such automatic evaluation methods are based on using LLMs as a judge, respectively auto-annotator. While several works on LLMs-as-a-judge showed that open-ended quality criteria show a high correlation between LLM-based judgments and human judgments~\cite{murugadoss2024evaluatingevaluatormeasuringllms}, works on correctness criteria and verification show the dependence on LLMs with high competence in the corresponding domain~\cite{lin2024criticbenchbenchmarkingllmscritiquecorrect}. In the absence of ground-truth data for example, we rely on models which can solve the task they are evaluating.

\paragraph{Dialogue systems}
The majority of the above-mentioned approaches are single turn approaches, synthetically generating a single user-AI interaction. Prior work on generating conversational data and dialogue acts with clarifications~\cite{deng-etal-2023-prompting, rahmani-etal-2023-survey, aliannejadi2021buildingevaluatingopendomaindialogue,budzianowski2020multiwozlargescalemultidomain} often leveraged human annotation with Crowdsourcing making data generation expensive. More recent approaches leverage LLMs to synthesize conversations based in grounded in knowledge sources~\cite{bao-etal-2023-synthetic} or hand-crafted templates~\cite{kulkarni-etal-2024-synthdst} for example. Especially generating conversations with clarification has been explored widely studied before in conversational recommendation~\cite{10.1145/3404835.3462839}, information retrieval~\cite{10.1145/3366423.3380126} , questions answering~\cite{guo2021abgcoqa} 
or for open-domain questions by LLMs~\cite{zhang2025llmsdesigngoodquestions}.

\paragraph{Data-centric tasks}
Related work on data-centric dialogues focus often on clarification of the user queries and regard the data as fixed~\cite{deng-etal-2022-pacific, aliannejadi-etal-2021-building, wu-etal-2023-inscit, guo2021abgcoqa}, we take a novel approach and assume the data is owned by the user, and can be collaboratively improved to meet a user's information need. Further, in this work we generate entire dialogues from existing target domain datasets, alleviating the burden of manually collecting conversations as in prior work like~\cite{deng-etal-2022-pacific}.

Following from prior success in leveraging code generation to solve data-centric tasks \citep{chen2021finqa, barke-etal-2024-solving}, we employ code generation with execution  to generate an answer for the table QA tasks. We generate code for two purposes: (1) several of the table QA tasks needed computations based on the content rather than look-ups; and (2) as part of the verification, helping to judged the solutions and identify important information from the task definition for answer generation. 



\section{Framework}
Our framework consists of two key components: 1) \textbf{a teacher-student method} to generate a curriculum for a student model (an LLM acting as table QA assistant) with clarification questions and corrections verified to lead to a correction of a wrong answer. We verify that the generated clarification questions by the student are useful in correcting wrong solutions and that the student can recover from its initial error to come to the final correct answer. 2) \textbf{a benchmarking method} to evaluate the student model on its ability to generate clarification questions and fix incorrect solutions on the synthetically generated curriculum without supervision by the teacher.






\subsection{Student-teacher method}
For the curriculum, we synthetically generate realistic human-AI conversations for table QA as data-centric task with missing information and verified correctable solutions by either, clarifications initiated by the AI assistant, or corrections initiated by the user. By verifying that a student model can come to a correct final answer after a clarification question and a correction using the teacher model, our approach makes sure we do not create unsolvable tasks. 

The original table QA task, without the need of a clarification or correction, is to find an answer $a$, as the student response $r$ to a given prompt $P$ with the task: 
$$T = (q, t)$$ 
for a question $q$ about a table $t$ represented as columns $[c_1, \cdots, c_m]$ and each column $c_i$ consisting of values $[v_{1i}, \cdots, v_{ki}]$. Depending on the dataset, we expect to have
a groundtruth answer $a^*$.

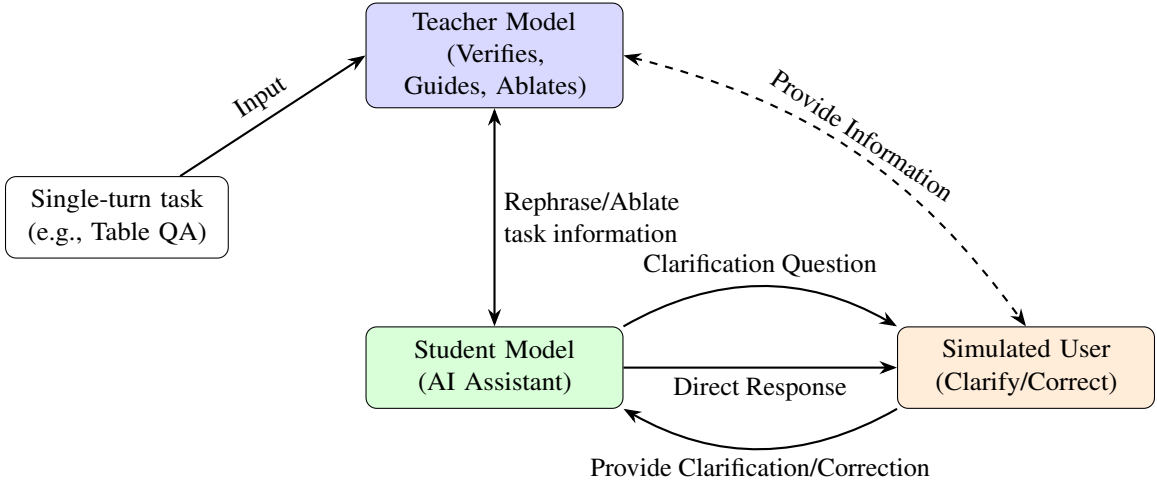
\begin{figure*}[ht]  
\centering  
\begin{tikzpicture}[
    scale=0.9, transform shape,
    node distance=1.5cm and 2cm,  
    >=Stealth,  
    box/.style={rectangle, draw, rounded corners, text width=3cm, align=center, minimum height=1.2cm},  
    teacher/.style={rectangle, draw, fill=blue!15, rounded corners, text width=3.5cm, align=center, minimum height=1.2cm},  
    student/.style={rectangle, draw, fill=green!15, rounded corners, text width=3.5cm, align=center, minimum height=1.2cm},  
    user/.style={rectangle, draw, fill=orange!15, rounded corners, text width=3.5cm, align=center, minimum height=1.2cm},  
    arrow/.style={->, thick}  
]  
  
\node[box] (initialtask) {Single-turn task \\ (e.g., Table QA)};  
\node[teacher, above right=1cm and 2cm of initialtask] (teacher) {Teacher Model \\ (Verifies, Guides, Ablates)};  
\node[student, below right=1cm and 2cm of initialtask] (student) {Student Model \\ (AI Assistant)};  
\node[user, right=4cm of student] (user) {Simulated User \\ (Clarify/Correct)};  
 
\draw[arrow] (initialtask) -- node[above,sloped]{Input} (teacher.west);  
\draw[arrow, <->] (teacher) -- node[right,align=left]{Rephrase/Ablate \\ task information} (student.north);  

\draw[arrow, dashed, bend left=20, <->] (teacher.east) to node[above, sloped, align=left]{Provide Information} (user.north);  

\draw[arrow] (student) -- node[below]{Direct Response} (user.west);  

\draw[arrow, bend left] (student.north east) to node[above]{Clarification Question} (user.north west);  
\draw[arrow, bend left] (user.south west) to node[below]{Provide Clarification/Correction} (student.south east);  
\end{tikzpicture}  
\caption{Teacher-Student framework illustrating synthetic dialogue generation with clarifications and corrections. Starting with an initial (single turn) table QA task, the teacher model rephrases and removes information such that the student cannot solve it (anymore). Then, the teacher guides the student how to ask clarification questions and how to use the provided user corrections to solve the task.}  
\label{fig:teacher_student_diagram}  
\end{figure*}

Starting from an initial task $T$, the teacher model checks whether the student model can solve the task by comparing the answer to the groundtruth $a^*$. If the student model can solve the task, they apply novel \textbf{ablation strategies} removing information $I$ from either the table $t$ or the question $q$, noted as $T_{-I}$. We verify that ablating this information makes the task unsolvable by the student model. Adding this information back in a follow-up clarification (initiated by the student model) or correction (initiated by a user), the task becomes solvable again for the student. To accomplish this, the teacher model guides the student model to generate \textbf{clarification} questions and perform a \textbf{correction}. The teacher model leverages the students' code solution to answer the question of the initial task to identify necessary information and for guidance to clarifications and corrections. See Fig.~\ref{fig:teacher_student_diagram}.

\paragraph{Ablation strategies}
To rephrase the original table QA tasks in such a way that a student model can only solve them using clarifications and corrections, we remove necessary information making the tasks not directly solvable. In contrast to previous approaches~\cite{deng-etal-2022-pacific} which only rephrase the user questions, we propose ablation strategies including both the user question and the table context.

The teacher model applies the strategies for ablating information $I$ from either the question or the table, noted as $T_{-I}$. Removing necessary information from the tables involves deleting either whole columns i.e. $I=c_i$, or values $I=v_{i,j}$ from the table which is needed to give the correct answer. Removing necessary information from the question is done by rephrasing the original question $q$ to $q'$ such that the question is either ambiguous or not answerable due to missing information, hence $I=\textit{diff}(q,q')$, the semantic difference between the two questions. In both cases, we use the teacher model to perform the ablation (see prompts in appendix).

If the removal of the necessary information makes the student model fail to produce the correct answer, producing an answer $a_{wrong}$, we add the new task $T_{-I}$ to a candidate set $C$ of potentially clarifiable and correctable questions under the student model, otherwise we disregard it.

\paragraph{Clarification}
After ablating the information, we instruct the student models to generate clarification questions and answer the unsolvable tasks.

Starting from the initial table QA task $T$ with ablated information $I$ in the candidate set $C$, a clarification is defined as: \newline
Given $T_{-I}$ and a candidate answer $a$, generate a clarification question $q_{cl}$ such that $I \subseteq r$ (the missing information is contained in the user's response) for $r$ a user response to $q_{cl}$; and find a new answer $a_{new}$ such that $a_{new} = a^*$.

To simulate conversations with clarification, the teacher model guides the student model to generate a clarification question $q_{cl}$ for each example in $C$, asking for more information. The teacher model will not give the groundtruth but gives hints of what information is needed to answer the question based on the above ablations.

Given the clarification question, the teacher model further simulates a user response $r$ containing the missing information. We verify that the clarification question together with the user response does help answering the questions by letting the student generate a new answer $a_{new}$, given the simulated response - evaluated by the teacher model comparing the new answer with the groundtruth. If the answer is correct, we add all to the set of clarifiable questions $Cl$.

\paragraph{Corrections}
Analogue to the clarifications, we instruct the student model to answer the unsolvable task with a given user correction without initiating a clarification question.

Starting from the initial table QA task $T$ with ablated information $I$ in the candidate set $C$ and a wrong answer $a_{wrong}$, instead of a clarification question, a correction is defined as: \newline
Given $T_{-I}$, a candidate answer $a$, and a user response $r$ such that $I \subseteq r$, find a new answer $a_{new}$ such that $a_{new} = a^*$. Hence, a correction task is similar to the clarification only that the model does not initiate a clarification question and returns a wrong answer instead.

To simulate conversations with corrections, the teacher model generates for each example from the candidate set $C$ a user response $r$ with correction instructions containing the necessary information $I$. The student generates a new answer $a_{new}$, given the simulated response, which is evaluated by the teacher model comparing the new answer with the groundtruth. If the answer is correct, we add all to the set of correctable questions $Co$.

\section{Curriculum}
\begin{table*}[t]
    \centering
    \small
    \begin{tabular}{l|l||ll|ll|l||ll|ll}
 &  \multicolumn{5}{c|}{TaT-QA} & \multicolumn{5}{c}{Wiki-TQ} \\
 \hline
 & Original & \multicolumn{2}{c|}{Ablate table} & \multicolumn{2}{c|}{Ablate question} & Original & \multicolumn{2}{c|}{Ablate table} & \multicolumn{2}{c}{Ablate question} \\
 \hline
 Student & $\accorig$ & $\accclar$ & $\acccorr$ & $\accclar$ & $\acccorr$ & $\accorig$ &  $\accclar$ & $\acccorr$ & $\accclar$ & $\acccorr$ \\
 \hline
 
 GPT4-Turbo & 0.829 & 0.797 & 0.882 & 0.566 & 0.834 & 0.665 & 0.711 & 0.749 & 0.507 & 0.862 \\
  \hline
Llama3.1 70b & 0.571 & 0.737 & 0.786 & 0.62 & 0.716 & 0.637 & 0.555 & 0.665 & 0.483 & 0.747 \\
Llama3.1 8b & 0.234 & 0.351 & 0.439 & 0.329 & 0.351 & 0.295 & 0.299 & 0.417 & 0.355 & 0.553 \\
\hline
Qwen2.5 72b & 0.703 & 0.645 & 0.786 & 0.62 & 0.773 & 0.579 & 0.587 & 0.717 & 0.573 & 0.784 \\
Qwen2.5 7b & 0.438 & 0.473 & 0.581 & 0.442 & 0.487 & 0.3 & 0.417 & 0.467 & 0.377 & 0.587 \\
\hline

    \end{tabular}
    \caption{Performance of student models while creating the synthetic curriculum for table QA \textbf{with} teacher supervision. For each benchmark TaT-QA (left) and Wiki-TQ (right) we report: (i) $\accorig$, the baseline accuracy on the original table questions; (ii) $\accclar$, the accuracy on tasks derived from originally correct examples after ablating essential information (from the table or question) and a model-initiated clarification; and (iii) $\acccorr$, the accuracy when the missing information is provided via a user-initiated correction. These metrics illustrate that larger models not only achieve higher baseline accuracies but also recover more effectively when corrections are provided, compared to when they must generate clarification questions on their own.}
    \label{tab:dataset_stats_1}
\end{table*}

To create a curriculum with synthetic conversations containing AI-initiated clarifications and user-initiated corrections 
we apply the above introduced teacher-student approach on two state-of-the-art table QA benchmark datasets, \textbf{TaT-QA}~\cite{zhu-etal-2021-tat} and WikiTableQuestions (\textbf{Wiki-TQ})~\cite{pasupat-liang-2015-compositional}. Both datasets contain user questions for a give table and a groudtruth answer. Our framework is universal applicable to all such datasets, as long as it contains a question, a table and a groundtruth answer. We select from each dataset 1000 samples from the dev-splits as initial table QA tasks. We use GPT4-Turbo~\cite{openai2024gpt4technicalreport} as teacher model with access to the groundtruth data in all our experiments and use five different LLMs of different sizes and model families as student model acting as table QA assistant: \textbf{GPT4-Turbo} itself as student model as capable as the teacher. \textbf{Llama3.1}~\cite{grattafiori2024llama3herdmodels} both 8b as small and 70b as general purpose language models and \textbf{Qwen2.5}~\cite{qwen2025qwen25technicalreport} both 7b as small and 72b as larger models with stronger reasoning and coding skills.

Starting from the samples, the teacher creates the conversations for a curriculum with a student model. On each sample, the teacher will ablate information from the questions and the table, but will only add the sample to the curriculum if the student can come to a final correct answer after a clarification or correction (see above). This means, the size of a curriculum depends on the accuracy of the student model on the clarification and correction tasks under the supervision of the teacher.

In Tab.~\ref{tab:dataset_stats_1}, we report the results of the generated curricula for the student models: ($\accorig$) measures the accuracies of the different student models as table QA assistant on the initial tasks from TaT-QA and Wiki-TQ 
(without the ablations); based on the correctly answered initial tasks (in set $C$), ($\accclar$) measures the accuracies of the student models on the clarification tasks with supervision from the teacher model; and ($\acccorr$) measures the accuracies of the student models on the corrections tasks with supervision from the teacher model.

We observe that the bigger models (GPT4, Llama3.1 70b, and Qwen2.5 72b) perform the best on the initial tasks, 
resulting in a larger pool of candidates for their curriculum (high $Acc_{Or}$). These models are also better in solving
the clarifications tasks (high $Acc_{Cl}$) and correction tasks (high $Acc_{Co}$) after the teachers ablates information. 
For example, Qwen2.5 72b can solve $70\%$ of the original TaT-QA tasks. For these tasks, we perform the above-described 
ablations such that Qwen2.5 72b can no longer solve them. Now, generating a clarification question, 
the model can correct the answer in $64\%$ of the cases; given directly the missing information as user correction, 
the model can correct the answer in $78\%$ of the cases. We see that generating a clarification question results in 
lower performance compared to when a user gives the correction directly (without making the model generate the 
clarification question). This discrepancy is due to the model not being able to integrate the information from the 
simulated user in their new solution. The simulated user response returns the correct missing information with high 
accuracy as discussed in the next section.

Based on the performance of the students, we can generate a curriculum with conversations containing a set $Cl$ of clarifications
and a set $Co$ of corrections. Starting from the initial table QA tasks, the sizes of these sets depend on the accuracy of the
corresponding student models, i.e., for Llama3.1 8b starting with $1000$ of the Wiki-TQ table QA tasks, the curriculum will contain
around $299$ clarifications and $417$ corrections. While this means we throw away a lot of table QA tasks due to the student not being able to finally get to a correct answer, we ensure the curriculum contains only tasks which the student potentially can solve by a clarification or a correction.


\subsection{Teacher Model Analysis}
We use the teacher model at several locations in our framework including a) measuring the correctness of a student
model answer, b) rephrasing the table QA task; and c) simulating a user response providing information to correct
an answer. We analyze the quality of the teacher on these tasks using both human annotations and automatic comparisons 
methods.

\paragraph{Teacher Model}
\begin{table}[t]
    \centering
    \small
    \begin{tabular}{l|ccccc}
    Teacher& GPT4 & \multicolumn{2}{c}{Llama3.1} & \multicolumn{2}{c}{Qwen2.5} \\
     Student & -Turbo &  70b &  8b & 72b & 7b \\
     \hline
      GPT4-Turbo &  0.86 & 0.64 & 0.66 & 0.81 & 0.63 \\
      Llama3.1 70b & 0.88 & 0.76 & 0.71 & 0.89 & 0.73 \\
      Llama3.1 8b &  0.78 & 0.57 & 0.56 & 0.766 & 0.56 \\
    \end{tabular}
    \caption{Alignment to human judgments on correctness \textbf{with} access to the groundtruth data. The Pearson correlations between our human annotations and correctness judgments of a teacher model, higher values indicate higher agreement on correctness.}
    \label{tab:critique_eval_groundtruth}
\end{table}


We conducted a human annotation of the answers from three different student models 
(GPT4-Turbo, Llama3.1 70b and Llama3.1 8b) for correctness and compared the results with judgments 
from different teacher models (all considered models: GPT4-Turbo, Llama3.1 70b and 8b, Qwen2.5 72b and 7b). 

We observe a strong correlation between our human annotations and correctness judgments of the student answers with GPT4-Turbo as well as Qwen2.5 72b as teacher model. These models also perform well on the original table QA tasks, which aligns with findings that the LLM's abilities on evaluation tasks depends on their ability to solve the task themselves~\cite{lin2024criticbenchbenchmarkingllmscritiquecorrect}. The major source of disagreements come from formatting issues i.e., rounding a number like $0.59$ given a groundtruth answer of $0.6$. Based on these finding we choose GPT4-Turbo as teacher model for all our experiments.

\paragraph{Task ablation}
To study the quality of ablating question information by the teacher model, we compare the resulting rephrased questions 
$q'$ to the original question, as well as to the corresponding human rephrased questions from the Pacific~\cite{deng-etal-2022-pacific} 
dataset as reference. We find that the rephrased questions in both approaches, ours from the teacher model and human 
ones from the Pacific dataset, are both semantically and lexically similar.  In Tab.\ref{tab:questions_simes}, we report 
the sentence similarity~\cite{reimers-2019-sentence-bert} as well as the 2-rouge scores~\cite{lin-2004-rouge} between 
original questions from TaT-QA, the rephrased questions by the teacher model and the questions from the Pacifc dataset. 

We find that the teacher rephrases the questions similar to how humans perform this task in the Pacific dataset. For example
one strategy to rephrase the question is to remove entities mentioned at the end, 
which is also observed in the Pacific dataset, i.e., the original question \textit{What was the discount rate for 2019?} from the TaT-QA dataset is rephrased both by our teacher model approach and the human annotators in the Pacific dataset as \textit{What was the discount rate?}.

\begin{table}[t]
    \centering
    \small
    \begin{tabular}{c|ccc|ccc}
        & \multicolumn{3}{c}{Sentence Sim} & \multicolumn{3}{c}{2-Rouge} \\
        & Orig & Ours & Pacific & Orig & Ours & Pacific \\
        \hline
        Ours & 0.77 & 1 & 0.74 & 0.71 & 1 & 0.71 \\
        Pacific & 0.84 & 0.74 & 1 & 0.77 & 0.71 & 1 \\
    \end{tabular}
    \caption{Average sentence similarity and 2-gram overlap (Rouge-score) between the original user questions in the TaT-QA dataset and their rephrased version to remove 
    information by both, our proposed teacher model and the human rephrases in the Pacific dataset.}
    \label{tab:questions_simes}
\end{table}

Analogue to the questions, we study the quality of ablating table information by measuring the differences to the
original table. In $98\%$ of all cases, the teacher changes the table after being instructed to remove relevant information. $36\%$ of the time the teacher removes columns, $67\%$ of the time the teacher removes values (not a whole column). We find that on average $5\%$ of the original table is removed (the main table stays intact) and in $85\%$ of the cases the table ablation results in a deficiency to solve the task, but at least one of the student models can clarify or correct it.

\paragraph{User responses}
We analyze how well the teacher model simulates a user response by measuring the amount of the ablated information 
contained to correct an answer. We perform a semi-automatic annotation where we first check how much of the ablated 
information (either from the question or the table) is contained in the user response. To account for rephrasing and 
simple misses due to formatting, we add an additional human annotation to perform the same task.

For ablating question information, we find that in more than $99\%$ of all simulated user responses, the ablated 
information is contained, either directly or rephrased. This is also independent on whether the student can finally 
answer the question. For ablating table information, we find that in $95\%$ of all simulate user responses, the response 
contains a subset of the ablated information from the table. Overall, in average $83\%$ of the ablated information is returned to  the student by the simulated user, but not $100\%$. We observe that the teacher model does not always return all the information it has removed from the table, but often only the amount needed to produce the correct answer. Looking at successful clarifications, we observe that the amount of table information in the user responses is lower compared to the unsuccessful cases. This is attributed too the model asking to general clarification questions.

\section{Evaluation}
\begin{table*}[t]
    \centering
    \small
    \begin{tabular}{l|lllll|lllll}
 &  \multicolumn{5}{c|}{TaT-QA} & \multicolumn{5}{c}{Wiki-TQ} \\
 \hline
Student & $P$ & $R$ & $F1$ &  $\accclar$ & $\acccorr$ & $P$ & $R$ & $F1$ &  $\accclar$ & $\acccorr$ \\
\hline
GPT4-Turbo & 0.82 & 0.36 & 0.5 & 0.81 &  0.91 & 0.81 & 0.38 & 0.51 & 0.67 & 0.75 \\
 \hline
Llama3.1 70b & 0.55 & 0.94 & 0.69 & 0.86 & 0.89 & 0.56 & 0.91 & 0.69 & 0.63 & 0.71\\
Llama3.1 8b & 0.29 & 1.0 & 0.45 & 0.63  & * & 0.59 & 1.0 & 0.74 & 0.57 & *\\
\hline
Qwen2.5 72b & 0.93 & 0.16 & 0.27 & 0.86 & 0.91 & 0.74 & 0.16 & 0.27 & 0.55 & 0.69 \\
Qwen2.5 7b & 0.44 & 1.0 & 0.61 & 0.78 & * & 0.45 & 0.98 & 0.62 & 0.71 & 0.6 \\
    \end{tabular}
    \caption{Performance of the student models on their created synthetic curriculum for table QA \textbf{without} teacher supervision using a \textbf{follow-up instruction prompt}. For each benchmark TaT-QA (left) and Wiki-TQ (right) we report: (i) precision, recall and F1-score on model-initiated clarification questions; (ii) $\accclar$, the accuracy on the examples with the ablated essential information after the model-initiated clarification; and (iii) $\acccorr$, the accuracy on the examples with the ablated essential information after a user-initiated correction. *No user-initiated corrections since the model asked always a clarification question. These metrics illustrate that smaller models almost always ask for clarification if they are prompted for such a decision. }
    \label{tab:instruct_student_tests_testqa}
\end{table*}

Based on the curricula generated for each student model from the dev-splits from TaT-QA and Wiki-TA as introduces above, we evaluate the student models on solving the tasks in $Cl$ and $Co$ \textbf{without} the help of the teacher. We still use the teacher model to evaluate the final answer and to simulate a user response, but the teacher no longer helps the student with asking the clarification question and with how to correct the wrong answer.

We evaluate the student using two prompting strategies: 1) Similar to the teacher-student interactions to generate a curriculum, we first prompt the model to answer the table question and then prompt the model as follow-up to decide whether to ask a clarification question or not. After the response from the student model, we simulate a user response either providing the information asked for in the clarification question or provide a direct user correction with the missing information if the model does not initiate clarification question. 2) We prompt the student model to answer the table questions and to decide whether to return either the answer or a clarification question in the same prompt, with zeroshot and fewshot prompting~\cite{brown2020languagemodelsfewshotlearners}. Additionally, we create a train curriculum based on the train-split of the table QA benchmarks TaT-QA and WikiTQ for the student models to measure whether they can improve their performance on the clarification and corrections tasks by finetuning. 

Given the curriculum for each student we measure precision, recall and F1-score on generating clarification questions based on the successful clarifications (positive class) and successful original tasks which did not need a clarification to answer (negative class). Further we measure the accuracy of the final answer after either the student initiated a clarification ($Cl$) or the user directed corrected the model with the missing information if the model did not initiate a clarification questions ($Co$).

\paragraph{1) Follow-up instruction prompt}
First we evaluate the student models to solve the clarification and corrections tasks when prompted as a follow-up after receiving an initial answer (Tab.~\ref{tab:instruct_student_tests_testqa}). 

We observe that the stronger models are more reluctant in generating clarification questions, resulting in a lower recall, but when they ask they are accurate, resulting in high precision i.e., Qwen2.5 72b has a low recall of $0.16$ but a high precision of $0.93$ on curriculum from TaT-QA. For the weaker models, we observe the opposite, a high recall and a low precision i.e., Llama3.1 8b has a recall of $1$ but a low precision of $0.29$. These models are less confident in their answers, asking almost always clarification questions, even if not needed. This behavior is consistent across the curriculum datasets. 

On both datasets, we see that recovery from a wrong answer without the teacher, with or without asking a clarification questions, is challenging even for highly capable student models like GPT4-Turbo. GPT4-Turbo for example achieves $81\%$ accuracy on tasks when it initiated a clarification, and $91\%$ when a user directly provided the correction. For the weaker models, like Llama3.1 8b, we observer even lower accuracies, ranging between $57\%$ and $67\%$. 

As observed before, we see higher accuracies on the corrections tasks ($Co$) when the student models decide not to ask a clarification question, compared to the clarification tasks ($Cl$) when the models ask a clarification question. For example, Llama3.1 70b is only able to give a correct final answer after clarification in $63\%$ of the tasks in the curriculum from Wiki-TQ, but in $71\%$ of the tasks without asking a clarification question.

\paragraph{2) Few-shot prompts}
\begin{table*}[t]
    \centering
    \small
    \begin{tabular}{l|lllll|lllll}
 &  \multicolumn{5}{c|}{TaT-QA} & \multicolumn{5}{c}{Wiki-TQ} \\
 \hline
Student & $P$ & $R$ & $F1$ &  $\accclar$ & $\acccorr$ & $P$ & $R$ & $F1$ &  $\accclar$ &  $\acccorr$ \\
\hline
 \multicolumn{11}{c}{Zeroshot}  \\
\hline
GPT4-Turbo & 0.98 & 0.31 & 0.47 & 0.9 & 0.86 & 0.98 & 0.44 & 0.61 & 0.65 & 0.79 \\
\hline
Llama3.1 70b & 0.96 & 0.09 & 0.16 & 0.8 & 0.75 & 0.93 & 0.22 & 0.35 & 0.53 & 0.76 \\
Llama3.1 8b & 0.5 & 0.01 & 0.03 & 0.67 & 0.46 & 0.62 & 0.03 & 0.05 & 0.55 & 0.61 \\
\hline
Qwen2.5 72b & 1.0 & 0.07 & 0.13 & 0.88 & 0.79 & 1.0 & 0.09 & 0.17 & 0.79 & 0.74 \\
Qwen2.5 7b & 0.92 & 0.22 & 0.36 & 0.55 & 0.66 & 0.84 & 0.34 & 0.49 & 0.57 & 0.72 \\
 \multicolumn{11}{c}{Fewshot}  \\
 \hline
GPT4-Turbo & 0.98 & 0.28 & 0.43 & 0.89 & 0.83 & 0.97 & 0.36 & 0.52 & 0.59 & 0.6 \\
\hline
Llama3.1 70b & 0.84 & 0.12 & 0.2 & 0.7 & 0.64 & 0.96 & 0.19 & 0.32 & 0.62 & 0.66 \\
Llama3.1 8b & 0.29 & 0.08 & 0.11 & 0.33 & 0.33 & 0.86 & 0.07 & 0.12 & 0.43 & 0.42 \\
\hline
Qwen2.5 72b & 1.0 & 0.1 & 0.19 & 0.85 & 0.74 & 0.91 & 0.08 & 0.14 & 0.71 & 0.71 \\
Qwen2.5 7b & 0.88 & 0.46 & 0.59 & 0.46 & 0.44 & 0.85 & 0.48 & 0.61 & 0.57 & 0.66 \\
 \multicolumn{11}{c}{Zeroshot - Finetuned}  \\
 \hline
 Llama3.1 8b & 0.31 & 0.89 & 0.45 & 0.59 & 0.55 & 0.57 & 0.53 & 0.55 & 0.5 & 0.43 \\
Qwen2.5 7b & 0.42 & 0.97 & 0.59 & 0.77 & 0.67 & 0.43 & 0.96 & 0.59 & 0.56 & 0.38 \\
    \end{tabular}
    \caption{Performance of the student models on their created synthetic curriculum for table QA \textbf{without} teacher supervision using \textbf{zero and fewshot prompts}. For each benchmark TaT-QA (left) and Wiki-TQ (right) we report: (i) precision, recall and F1-score on model-initiated clarification questions; (ii) $\accclar$, the accuracy on the examples with the ablated essential information after the model-initiated clarification; and (iii) $\acccorr$, the accuracy on the examples with the ablated essential information after a user-initiated correction. These metrics illustrate with zero- and fewshot prompts all models are reluctant in asking clarifications before returning an answer. Finetuning can increase the recall and also improve the accuracies on the final answers.}
    \label{tab:fewshots_student_tests_testqa}
\end{table*} 

In contrast to testing the student models with a follow-up prompt to decide whether to ask a clarification question, next we test them in a single step, prompting them only once to generate an answer and clarification question if needed. This tests the student models on their self-criticizing abilities. We are testing the students using zero- and fewshot prompting (Tab.~\ref{tab:fewshots_student_tests_testqa} - top and middle section). For the fewshot prompts we extract $4$ examples from the train curricula and present them in two different orders to avoid positional biases.

We observe that, in contrast to testing the students with a follow-up prompt, all models are now more reluctant to ask clarification questions. With zeroshot prompting for example, the smaller student models Llama3.1 8b, and Qwen2.5 7b drop in recall from $1$ to $0.22$, respectively $0.01$ when not prompted as a follow-up to decide to ask a clarification question and instead prompted to make this decision before returning an answer.

\paragraph{Finetuning}
Besides using the curricula with the synthetic conversations to benchmark the student models for their capabilities, we additionally  generate a train set from a curriculum generated from the train-splits of the table QA datasets (Tab.~\ref{tab:fewshots_student_tests_testqa} - bottom). We generate for both datasets, TaT-QA and Wiki-TQ, a train curriculum with a total of $2000$ successful clarifications and corrections. We finetune Llama3.1 8b and Qwen2.5 7b on these tasks and test the models with zeroshot prompting on the dev curricula, analogue to the previous evaluation.  We observe that after finetuning, the recall of the student models increase again, making the models asking clarification questions more often. This results in performance similar to when we use follow-up instruction prompts. Qwen2.5 7b benefits more from finetuning than Llama3.1 8b, improving not only in recall but also in accuracy after clarification.

\section{Conclusion}
In this work, we introduced a novel student–teacher framework to generate a curriculum of synthetic conversations with clarifications and corrections for table QA tasks. Our approach leverages a strong teacher model to verify that every synthesized dialogue constitutes a solvable sub-task, ensuring that the generated data provides a reliable benchmark for evaluating large language models.

We conducted extensive experiments using two well-known table QA datasets (TaT-QA and Wiki-TQ) and evaluated a range of student models, from highly capable systems such as GPT4-Turbo to smaller models like Llama3.1 8b and Qwen2.5 7b. Our results show that stronger models not only achieve higher baseline accuracy on original tasks but are also more effective at handling information deficiencies when guided by teacher feedback. Moreover, we demonstrate that finetuning on the synthetic curriculum significantly improves student models’ ability to generate clarification questions and correctly incorporate user-initiated corrections.

\section{Limitations}
The curricula are dependent on the characteristics of a specific student model, meaning it may overlook failure cases that are common across other models. Additionally, the framework requires a strong teacher LLM to generate and validate the synthetic conversations—although this is still more cost-effective than human annotation, it can be expensive and the teacher model may itself exhibit issues such as hallucinations, biases, or lower performance than human evaluators.

\bibliography{custom}

\appendix

\section{Experiment setups and resources}
In all our setups we use GPT4-Turbo (2024-04-09) as teacher model with a temperature of $0$. As student models with use GPT4-Turbo (2024-04-09), Llama3.1-Instruct 8b and 70b, Qwen2.5-Instruct 7b and 72b from Huggingface with temperature of $0$. For Llama3.1 and Qwen2.5 inference and finetuning, we use a single A100 GPU with 80GB RAM. For finetuning we use LORA optimization with a rank and $\alpha$ of $16$, running for $4$ epochs with a learning rate of $1e-4$.

\label{Next_curriculum}
\section{Evaluation on Next Curriculum}
\begin{table*}[t!]
    \centering
    \small
    \begin{tabular}{l|llll|l|llll|l}
 &  \multicolumn{5}{c|}{TaT-QA} & \multicolumn{5}{c}{Wiki-TQ} \\
 \hline
Student & $P$ & $R$ & $F1$ &  $\accclar$ & $\acccorr$ & $P$ & $R$ & $F1$ &  $\accclar$ & $\acccorr$ \\
\hline
 \multicolumn{11}{c}{Zeroshot}  \\
 \hline
 \multicolumn{11}{c}{Llama3.1 70b curriculum}  \\
 \hline
Llama3.1 8b & 0.85 & 0.2 & 0.33 & 0.27 & 0.25 & 0.65 & 0.16 & 0.26 & 0.19 & 0.13 \\
 \multicolumn{11}{c}{Qwen2.5 72b curriculum}  \\
 \hline
Qwen2.5 7b & 0.74 & 0.5 & 0.6 & 0.28 & 0.36 & 0.89 & 0.52 & 0.66 & 0.23 & 0.25 \\
\hline
\multicolumn{11}{c}{} \\ 
 \multicolumn{11}{c}{Zeroshot - Finetuned}  \\
 \hline
 \multicolumn{11}{c}{Llama3.1 70b curriculum}  \\
 \hline
Llama3.1 8b & 0.55 & 0.99 & 0.71 & 0.34 & 0.19 & 0.51 & 0.66 & 0.58 & 0.3 & 0.31 \\
\hline
  \multicolumn{11}{c}{Qwen2.5 72b curriculum}  \\
 \hline
Qwen2.5 7b & 0.52 & 0.94 & 0.67 & 0.68 & 0.59 & 0.34 & 0.89 & 0.49 & 0.4 & 0.26 \\
\hline
    \end{tabular}
    \caption{Zero- and Fewshot prompt results for TaT-QA for the corrections ($Co$) and clarification ($Cl$) tasks from the curriculum of the next larger model in the same family for Llama3.1 and Qwen2.5.}
    \label{tab:next_curriculum_student_tests_testqa}
\end{table*}

Additionally, we test two student models for their performance on the curriculum from the next larger models of the same family, Llama3.1 8b on the curriculum from Llama3.1 70b and Qwen2.5 7b on the curriculum from Qwen2.5 72b, see results in Tab.\ref{tab:next_curriculum_student_tests_testqa}.

We observe that the student models ask more often clarification questions on the next larger student models' curriculum compared to both their performance on their own curriculum and the next larger model on their corresponding curriculum~\ref{tab:fewshots_student_tests_testqa}. For example, Qwen2.5 7b has a recall of $0.22$ on their own curriculum and Qwen 2.5 72b has a recall of $0.07$ on their own curriculum, while Qwen2.5 7b has a higher recall of $0.5$ on the curriculum of Qwen2.5 72b. On the other hand, we see that the smaller student models are less capable of given the final correct answer on the curriculum of the next larger model i.e. Llama3.1 8b has only an accuracy of $0.19$ on the clarification tasks from the Llama3.1 70b curriculum, while gets $0.53$. Finally, finetuning the student models as described on the last section, results in both higher recall and higher accuracy.

\section{Prompts}

\begin{figure*}
\begin{tcolorbox}[colback=gray!10, colframe=black, width=\textwidth, title=Teacher Prompt: Judge the student model answer.]
\begin{Verbatim}[fontsize=\footnotesize, breaklines=true]
#Instruction#
# ########## #
Review the provided answer for a user question. Determine whether it is correct and answers the user request properly. 
If wrong, find the problems with it. Finally, conclude with '[[correct]]' if the provided model answer is correct or 
'[[wrong]]' if it is incorrect.

# ########## #
#Task#

## User request was: 
{question}

## Input table:
{table}

## The model answer is:
{output}

## The correct answer is:
{solution}

Review the model answer and compare to the correct answer in order to decide the verdict of [[correct]] or [[wrong]].
If the model answers the question correctly, compared to the provided the correct answer, mark the model's answer as [[correct]].
If the model answer gives an incorrect answer, an execution failure, or otherwise avoids giving an answer, mark the model answer as [[wrong]].

Give a step-by-step explanation including requirements and an analysis.
Provide your analysis and verdict in JSON format as follows: 
```json
{"analysis": "your analysis", "verdict": "[[correct]], [[wrong]]"}
```
\end{Verbatim}
    \end{tcolorbox}
    \caption{Prompt for the teacher model to judge a students answers for table QA tasks.}
    \label{fig:prompt_teach_judges}
\end{figure*}

\begin{figure*}
    \begin{tcolorbox}[colback=gray!10, colframe=black, width=\textwidth, title=Teacher Prompt: Ablate information from the question]
    \begin{Verbatim}[fontsize=\footnotesize, breaklines=true]
# I have the following code solution for this question:

# Question:
{question}

# This is the solution:
{solution}

Look at the solution and identify two necessary pieces of information from
the question above that are needed to correctly compute the solution. Each information
piece needs to be part of the question and important to answer it, so that when
removing it from the question, you can cannot answer it.
Return these as:
```json
[{"piece 1": "description"}, {"piece 2": "description"}]
```
\end{Verbatim}
\end{tcolorbox}
\caption{Prompt for the teacher model to ablate information from the user question.}
\label{fig:prompt_teacher_question_ablate}
\end{figure*}

\begin{figure*}
    \begin{tcolorbox}[colback=gray!10, colframe=black, width=\textwidth, title=Teacher Prompt: Ablate information from the table]
    \begin{Verbatim}[fontsize=\footnotesize, breaklines=true]
# I have the following code solution for this problem:

Table:
{table}

Query:
{question}

# This is the solution:
{solution}


Look at the solution code and identify two necessary pieces of information from
the input table that are needed to correctly compute the solution. Return these as:
```json
[{"piece 1": "description"}, {"piece 2": "description"}, {"piece 3": "description"}]
```
\end{Verbatim}
\end{tcolorbox}
\caption{Prompt for the teacher model to ablate information from the table.}
\label{fig:prompt_teacher_table_ablate}
\end{figure*}

\end{document}